\documentclass[letterpaper, 10 pt, conference]{ieeeconf}  % Comment this line out if you need a4paper

\IEEEoverridecommandlockouts                            
\usepackage{gensymb}
\usepackage{amsmath}
\usepackage{graphicx}
\usepackage{multirow}
\usepackage{subfigure}
\usepackage{hyperref}
\usepackage{flushend}

\usepackage{caption}

\usepackage{color, colortbl}
\definecolor{Gray}{gray}{0.9}
\usepackage[noadjust]{cite}
\usepackage[font={footnotesize}]{caption}

\title{\LARGE \bf
CRITERIA: a New Benchmarking Paradigm for Evaluating Trajectory Prediction Models for Autonomous Driving 
}

\author{Changhe Chen$^{1}$, Mozhgan Pourkeshavarz$^{2*}$, Amir Rasouli$^{2}$\\
\thanks{$^{1}$University of Toronto. Work done during internship at Huawei.}
\thanks{$^{2}$Noah's Ark Laboratory, Huawei Technologies Canada.}
\thanks{*Corresponding author {\tt\footnotesize Mozhgan.Pourkeshavarz@huawei.com}}
}

\begin{document}

\maketitle
\thispagestyle{empty}
\pagestyle{empty}

%%%%%%%%%%%%%%%%%%%%%%%%%%%%%%%%%%%%%%%%%%%%%%%%%%%%%%%%%%%%%%%%%%%%%%%%%%%%%%%%
\begin{abstract}
Benchmarking is a common method for evaluating trajectory prediction models for autonomous driving. Existing benchmarks rely on datasets, which are biased towards more common scenarios, such as cruising, and distance-based metrics that are computed by averaging over all scenarios. Following such a regiment provides a little insight into the properties of the models both in terms of how well they can handle different scenarios and how admissible and diverse their outputs are. There exist a number of complementary metrics designed to measure the admissibility and diversity of trajectories, however, they suffer from biases, such as length of trajectories.

In this paper, we propose a new benChmarking paRadIgm for evaluaTing trajEctoRy predIction Approaches (CRITERIA). Particularly, we propose 1) a method for extracting driving scenarios at varying levels of specificity according to the structure of the roads, models' performance, and data properties for fine-grained ranking of prediction models; 2) A set of new bias-free metrics for measuring diversity, by incorporating the characteristics of a given scenario, and admissibility, by considering the structure of roads and kinematic compliancy, motivated by real-world driving constraints; 3) Using the proposed benchmark, we conduct extensive experimentation on a representative set of the prediction models using the large scale Argoverse dataset. We show that the proposed benchmark can produce a more accurate ranking of the models and serve as a means of characterizing their behavior. We further present ablation studies to highlight contributions of different elements that are used to compute the proposed metrics\footnote{Code available at \url{https://github.com/huawei-noah/SMARTS/tree/CRITERIA-latest/papers/CRITERIA}}.
\end{abstract}

\section{Introduction}
Trajectory prediction is one of the main components of autonomous driving systems for estimating the behavior of other road users for safe motion planning. One of the challenges in prediction is that the future behaviors of road users are uncertain due to various hidden factors, such as unknown intentions. Thus, trajectory prediction models typically produce multimodal outputs to capture all possibilities.

 Existing trajectory prediction benchmarks \cite{Argoverse, sun2020scalability, wilson2023argoverse} consist of driving data and distance-based metrics to evaluate the accuracy of the models against the ground truth. The datasets, however,  are typically biased towards more common scenarios, such as cruising. Consequently, when metrics are averaged over the entire data, the benchmark tends to favor the models that perform better on common cases and provides little insight into the differences among the models and how they behave in rarer situations. This would limit the ability to observe the effects of design choices on particular driving scenarios.

\begin{figure}
\vspace{0.2cm}
    \centering
\includegraphics[width=0.47\textwidth,trim={0cm 0cm 10cm 5cm},clip]{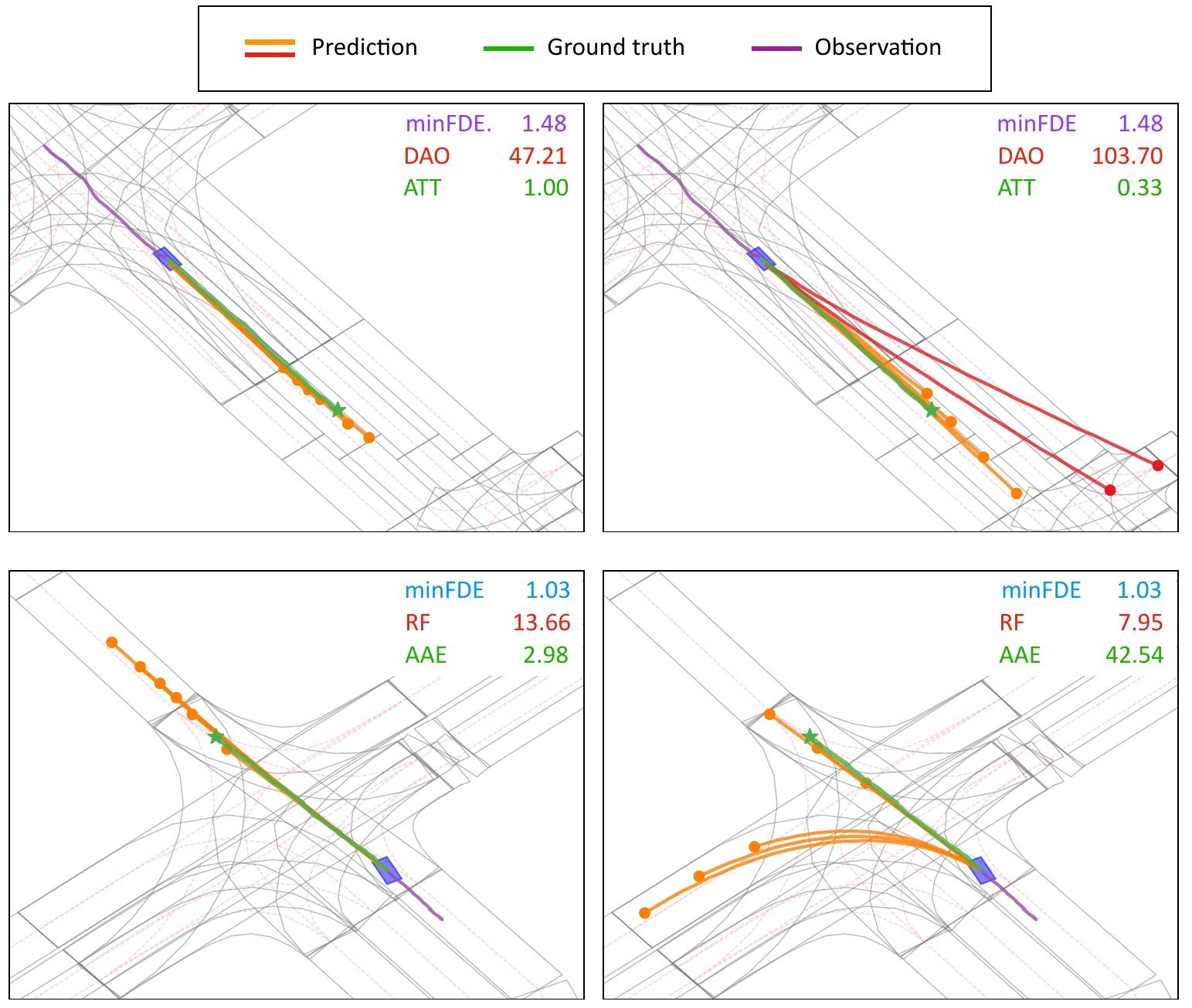}
    \caption{Examples of two scenarios with two different prediction sets where diversity and admissibility of the trajectories are not reflected in minimum final displacement error (minFDE). In the top two predictions, minFDEs are the same, even though in the right prediction, the red trajectories are inadmissible. In the bottom row, minFDE of both cases are the same, but the right example has a more diverse trajectory.}
    \label{fig:intro_fig}
    \vspace{-3mm}
\end{figure}

Another shortcoming of existing benchmarks is the over-reliance on distance-based metrics as they are insufficient for assessing the diversity and admissibility of generated trajectories. For the former, one requires to measure the relative positioning of multimodal outputs, therefore the diversity is not effectively captured by measuring error against single-biased ground truths. For admissibility, understanding of the road structure and dynamical constraints is important. Neither of these factors, however, are adequately perceivable when merely computing distance error. The aforementioned shortcomings point to the need for a better measure of goodness of predicted trajectories (see Fig. \ref{fig:intro_fig}).

In the literature, there are a number of metrics proposed for measuring diversity and admissibility of trajectories. However, the existing diversity-based metrics \cite{park2020diverse,ma2021likelihood} are based on trajectory end-points, therefore they do not distinguish between lateral (following different routes) and longitudinal (different driving behavior, e.g. speed up/down) diversities, which have different meanings in real-world driving. Furthermore, existing admissibility metrics \cite{chang2019argoverse, park2020diverse} are heavily biased towards the length of the trajectories, meaning that the longer the generated trajectories are the better metric values get. Additionally, since the entire drivable area is considered valid by these metrics, they verify trajectories even though they fall on invalid parts of the map, such as lanes going in the opposite direction of the vehicle (see Fig. \ref{fig:intro_fig}).

To this end, we propose a new benChmarking paRadIgm for evaluaTing trajEctoRy predIction Approaches (CRITERIA) with a goal of providing a more accurate characterization of trajectory prediction models, and consequently better design choices. We propose a new approach for extracting different driving scenarios. In detail, we categorize driving scenarios on different levels of specificity based on agreement among models' performance as well as characteristics of the scene and driving task. This would provide a fine-grained categorized scenarios for better understanding of models' behavior under different conditions. 

Furthermore, we propose a novel set of bias-free metrics for accurately measuring the performance of trajectory prediction models. Our metrics present a new insight into the diversity and admissibility of trajectories by defining road and kinematic compliancy for normal driving behavior, which are vital aspects of real-world driving. Our metrics help characterize models' behavior more effectively and provide a different, yet more accurate ranking scheme than the existing metrics. 

In summary, our \textbf{main contributions} are as follows: We present a new benchmarking paradigm named CRITERIA, to evaluate trajectory prediction models. For this purpose, 1) we present a new scheme that relies on agreement among prediction models as well as driving scenes and tasks to extract driving scenarios with different levels of specificity. 2) We propose a new set of metrics to assess diversity and admissibility of trajectories while mitigating biases in the existing metrics with the ability to define conformity at different levels, including road and kinematic compliancy. 3) We conduct extensive experimentation to highlight the effectiveness of the proposed benchmarking paradigm for ranking and characterizing prediction models. 4) Via ablation studies, we show the contribution of different elements of our admissibility metric and their corresponding values on overall ranking of the models.

\section{Related work} \label{Section2}
\subsection{Trajectory Prediction}
An extensive body of literature is dedicated to trajectory prediction in the context of autonomous driving \cite{nayakanti2023wayformer, pini2023safe, cui2023gorela,wang2023ganet,su2023uncertainty,gu2023vip3d,ivanovic2023expanding, zhou2023query, jain2023ground}. Existing methods investigate various representations and approaches to learning scene context, including rasterized images \cite{chai2019multipath,salzmann2020trajectron++,gilles2022gohome}, point-clouds \cite{ye2021tpcn} processed with convolutional neural networks, and vectorized representations processed using different architectures, such as graph neural networks \cite{gao2020vectornet,mercat2020multi,gu2021densetnt,zeng2021lanercnn} or transformer architectures\cite{huang2022multi, zhou2022hivt, girgislatent2022,liu2021multimodal} often combined with sophisticated fusion mechanisms \cite{liang2020learning}. To address the inherent uncertainty in predicting future trajectories, many models resort to generating multimodal trajectories in order to cover the space of possibilities. There are different approaches to achieving this goal, including the use of generative adversarial networks \cite{sadeghian2019sophie} \cite{zhao2019multi}, conditional variational autoencoders \cite{sohn2015learning}, or training sampling networks \cite{bae2022non, ma2021likelihood}. Some methods also use trajectory endpoints \cite{gu2021densetnt} \cite{zhao2021tnt} or targets \cite{girase2021loki, gu2021densetnt, mangalam2021goals, zhao2021tnt} to model possible future intentions of vehicles to encourage more diverse predictions. Regardless of the method of choice, there is a need for an effective mechanism to evaluate the quality of generated trajectories both in terms of diversity and admissibility, i.e. compliancy to road structure or dynamical limitations.

\subsection{Evaluation Metrics}
\subsubsection{Distance-based Metrics} 
Distance-based metrics are the most commonly used metrics in the domain of trajectory prediction. They are often computed based on the Euclidean distance between the ground truth and the generated trajectories and reported as average displacement error (ADE) or/and final displacement error (FDE) \cite{Argoverse}. In the multimodal setting, the mode with a minimum error (e.g. minADE, minFDE) with respect to the ground truth is selected as the reported value. An extended version of these metrics also accounts for the probability of the selected mode, taking the form of brier-minADE/FDE as in \cite{Argoverse}. Miss Rate (MR) \cite{yeh2019diverse} \cite{lee2017desire} is another common metric in this category that corresponds to the proportion of trajectories whose Euclidean prediction error surpasses a predetermined threshold. Mean average Precision (mAP) \cite{ettinger2021large} is a variation of MR that also considers positional uncertainty in predictions.

\subsubsection{Admissibility Metrics} Admissibility metrics determine whether trajectories satisfy map constraints, e.g. road boundaries. Some of the metrics in this category include Driveable Area Compliance (DAC) and off-road rate \cite{Argoverse}, which measure the ratio of trajectories that fall within the driveable area. Both of these metrics, however, consider the entire scene as the drivable area and do not penalize the model if the predictions go outside of the valid lanes, e.g. the ones in the opposite direction. Another metric is referred to as Driveable Area Occupancy (DAO) \cite{park2020diverse}, which measures the proportion of pixels occupied by the trajectories in the drivable area. Similar to other metrics, DAO is highly biased towards the length of trajectories and it does not constrain the extent of the drivable area. This means, the longer trajectories get the higher (better) the value of the metric.

\subsubsection{Diversity Metrics} 
As the name implies, these metrics are designed to measure how well the generated trajectories cover future possibilities. Existing diversity metrics, such as Ratio of avgFDE to minFDE (RF) \cite{park2020diverse}, minimum average self-distance (minASD), and minimum final self-distance (minFSD) \cite{ma2021likelihood} depend on the distances between generated trajectories. For instance, RF measures the spread of predicted trajectories using their final positions \cite{park2020diverse}, which is highly dependent on the single-biased ground truth. MinASD and minFSD are calculated using the minimum average and final distance between all pairs of predicted trajectories \cite{ma2021likelihood}. This means that, first, these metrics do not distinguish between lateral and longitudinal diversity, which is fundamentally different in real-world driving, and second, they are biased towards the length of the trajectories.

\section{CRITERIA}
Characterizing existing models and analyzing their behavior under different conditions can provide new insights for future research directions and help improve trajectory prediction models. In this regard, we propose a new benChmarking paRadIgm for evaluaTing trajEctoRy predIction Approaches (CRITERIA). More superficially, we present a new method for extracting scenarios in a systematic and meaningful way (Section \ref{eval-proto}) and a set of novel metrics for measuring diversity and admissibility of trajectory prediction models (Sections \ref{adm-sec} and \ref{div-sec}).

\subsection{Problem Formulation}
Trajectory prediction task can be viewed as an optimization problem. At time step $t$, let consider the past trajectory of the $i$-th agent as a set of 2D coordinates in bird's eye view over some observation horizon $L$ time steps $X_i = \{(x_i, y_i)^{t-L+1}, \cdots,(x_i, y_i)^{t}\}$. Then, the objective is to predict future trajectories $Y_i = \{(x_i, y_i)^{t+1}, \cdots,(x_i, y_i)^{t+T}\}$, where $T$ is the prediction horizon. In this formulation, additional contextual information, such as road lane boundaries and links are used. In the multimodal setting, prediction models output $K$ different trajectories (referred to as modes) per each vehicle. For simplicity, in the remainder of this paper, we refer to $t-L+1$ and $t+T$ as $L$ and $T$, respectively.

\subsection{Scenario Extraction} \label{eval-proto}
Autonomous driving datasets consist of a variety of scenarios with different levels of complexity, such as simple cruising and turning at intersections. These datasets, however, are biased towards more common scenarios, such as cruising. As a result, evaluating prediction models by averaging the performance over the entire datasets tells little about the strengths and weaknesses of the models and also favors the models that perform better on the common cases.

Trajectory prediction models have different architectural designs and often behave differently when applied to the same samples. For instance, the same models may generate different distributions, and trajectories with different lengths, curvature, etc. As a result, it is essential to extract driving scenarios in a meaningful way to better reflect such differences among the models. To this end, we propose a method to extract driving scenarios based on three criteria, namely the road structure, models' performance, and data properties.

\subsubsection{Road Structure} We select two criteria that impose two different driving behavior, namely straight roads that impose cruising behavior and intersections that require turning behavior. For each time step in the ground truth trajectories, we obtain all lane segments intersecting with the current route with a radius of $100$ meters from the current position of the vehicle. We tag the scenario as a turn if the adjacent lane segments contain at least one intersection with either a right or left turn, otherwise, it is considered a cruising scenario.

\subsubsection{Model Performance} We determine the difficulty of scenarios by aggregating models' accuracy computed based on the average minFDE of all models that are being evaluated on each scenario. We categorize the scenarios into three groups: hard, medium, and easy, with $\alpha_1$, $\alpha_2$, and $\alpha_3$ percent each, where we set them empirically. 

\subsubsection{Data Properties} The last factor we consider is the length of trajectories as they implicitly reflect different driving styles. For example, future trajectories indicate slowing down behavior to yield or stop at an intersection. We empirically identify a threshold value $\beta$ based on which we categorize trajectories into short and long groups.

\subsection{Admissibility Metric} \label{adm-sec}
We propose Admissibility Triad Test (ATT) metric comprised of three tests motivated by the admissibility definition in real-world driving scenarios. The trajectory is considered admissible if it passes all tests, otherwise, it is counted as inadmissible. The final metric is computed based on the ratio of admissible trajectories. The tests are described below.

\subsubsection{Road Boundary Compliancy Test} This test determines whether a trajectory is compliant to the road boundaries. For this purpose, we examine the positions of trajectory points at each time step (point-wise) and if a point falls within the drivable area, it passes the test.

\subsubsection{Road Boundary Alignment Test} This test identifies whether the trajectory orientation is aligned with the lane driving direction. Let consider some trajectory $\tau_i$. We first find the orientation of the trajectory based on its last three time steps. Then, we retrieve the lanes from the HD map where the trajectory's last three time steps are situated. Subsequently, we calculate the orientations $\tau_l$ of these lanes, and then calculate the orientation variances $\Delta \Theta$ between $\tau_l$ and $\tau_i$ for the most recent three time steps. Next, we obtain the confidence level between the trajectory's orientation and that of the lanes as below:

\begin{equation} 
 \label{LAC_conf}
\mathcal{C} = max(0, 1-\frac{\Delta \Theta}{\pi}).
\end{equation}
The maximum confidence value is then compared with $\text{threshold}_{\text{lac}}$. We set the $\text{threshold}_{\text{lac}} = 0.5$, which indicates that the orientation difference should be less than 90$\degree$. The trajectory passes the test if the maximum confidence of the final three time steps is more than $\text{threshold}_{\text{lac}}$, meaning that it is aligned with the road boundaries.

\subsubsection{Kinematic Compliancy Test} \label{Kinematic}
This test determines if a trajectory's longitudinal acceleration is within an admissible range. We begin by calculating the longitudinal acceleration based on the average initial acceleration and final acceleration of predicted trajectories. Then, by defining an admissible range for normal driving behaviors \cite{bae2022selfdriving}, the trajectories pass the test if their longitudinal acceleration falls within the admissible ranges. 

\subsection{Diversity metrics} \label{div-sec}
We consider diversity as the spread of predicted trajectories in lateral and longitudinal directions; hence, we propose two metrics to capture both aspects.

%%%AAD: Average Angle Discrepancy%%%
\subsubsection{AAE: Average Angular Expansion}
AAE determines lateral diversity by computing the differences between predicted trajectories' angles. It captures the lateral variation of trajectories independently of their lengths. Mathematically speaking, given $K$ trajectories in a multimodal setting, for each pair of trajectories $(i, j), i\neq j$, we calculate the angle between the first and last time steps as follows:
\begin{equation}
{AE_{i,j} = \angle\left[\vec{v}_i, \quad \vec{v}_j\right] }
\label{TAD1}
\end{equation}
where $\vec{v}$ is a trajectory vector comprised of the first and last points $\vec{v} =\langle \hat{x}^T - \hat{x}^t, \hat{y}^T -\hat{y}^t\rangle$ of each trajectory. The angle between two trajectories is calculated once $AE_{ij}=AE_{ji}$. Then, we compute AAE by averaging over all trajectory pairs' angular expansion.

%%%%ADD: Average Displacement Discrepancy. %%%
\subsubsection{AMV: Average Magnitude Variation}
The AMV metric measures longitudinal diversity based on the rate of change (speed-up/down driving behavior) of trajectories. In this manner, however, it is important to exclude scores resulted from exaggerated longitudinal changes, i.e. trajectories should be kinematically feasible besides contributing to the diversity. In this regard, we first run the kinematic compliancy test mentioned in Section \ref{Kinematic} and clip the trajectories at their maximum acceptable length according to the test. Then, we compute longitudinal diversity as follows: given $K$ trajectories in a multimodal setting, for each pair of trajectories $(i, j), i\neq j$, we calculate the magnitude of the difference between the two trajectories per each time step and then average over the entire prediction horizon $T$,
\begin{equation}
MV_{i,j}=\sum_{t=1}^{t=T}| \|\vec{v^{\prime}}_i^{t}\| - \|\vec{v^{\prime}}_j^{t}\|| 
\end{equation}
where $\|\vec{v^{\prime}}^t\| $ is the magnitude of the trajectory vector $\vec{v} =\langle \hat{x}^t - \hat{x}^{t-1}, \hat{y}^t -\hat{y}^{t-1}\rangle$ starting at time step $t-1$ and ending at $t$ that is passed by kinematic the compliancy test. The magnitude variation between two trajectories is calculated once $MV_{ij}=MV_{ji}$, and then AMV is calculated by taking an average over all trajectory pairs' magnitude variances.

\section{Experiments}

\subsection{Experimental Setup}
\textbf{Dataset.} We conduct our evaluations on the Argoverse dataset \cite{chang2019argoverse}, where the task is to predict 3 seconds of future trajectories given 2 seconds of past observations. This dataset consists of more than $300K$ real-world driving sequences, which are split into train (205K), validation (39K), and test (78K) sets without geographical overlap, along with corresponding HD maps.

\textbf{Metrics.} We report on accuracy metrics, minADE \cite{Argoverse} and minFDE \cite{Argoverse}. For diversity and admissibility, in addition to the proposed metrics AAE, AMV and ATT, we report on RF \cite{park2020diverse}, minASD/minFSD \cite{ma2021likelihood}, DAO \cite{park2020diverse}, and DAC \cite{Argoverse}.

\textbf{Models.}
We select the models with different architectural designs to better reflect the characteristics of each approach in different scenarios. These models are: \textit{TNT} \cite{zhao2021tnt} as a target-based model, \textit{LaneGCN} \cite{liang2020learning} as a graph-based method that uses road information, \textit{FTGN} \cite{aydemir2022trajectory} as a graph-based model that relies on the entire scene, \textit{MMTransformer} \cite{liu2021multimodal} as a fully transformer-based model, and finally \textit{HiVT} \cite{zhou2022hivt} as the recent SOTA model which utilizes an innovative combination of transformers and recurrent networks. For all models, except TNT\footnote{The implementation used is from \url{https://github.com/Henry1iu/TNT-Trajectory-Prediction}}, we use the official code released by the authors.

\textbf{Parameter setup.}
\begin{table}[]
\centering
\vspace{0.2cm}
\caption{Number of scenarios in each category, where C1, C2, and C3 indicate the road structure, model performance, and data characteristics criteria, respectively.}
\begin{tabular}{c|c|c|c}
\hline
C1  & C2  & C3 & Number \\ \hline
\multirow{4}{*}{Hard} & \multirow{2}{*}{Turn} & short & 2230 \\ \cline{3-4} 
 &  & long & 1124 \\ \cline{2-4} 
 & \multirow{2}{*}{Cruising} & short & 248 \\ \cline{3-4} 
 &  & long & 345 \\ \hline
\multirow{4}{*}{Middle} & \multirow{2}{*}{Turn} & short & 8424 \\ \cline{3-4} 
 &  & long & 4179 \\ \cline{2-4} 
 & \multirow{2}{*}{Cruising} & short & 1565 \\ \cline{3-4} 
 &  & long & 3594 \\ \hline
\multirow{4}{*}{Easy} & \multirow{2}{*}{Turn} & short & 5106 \\ \cline{3-4} 
 &  & long & 4581 \\ \cline{2-4} 
 & \multirow{2}{*}{Cruising} & short & 2544 \\ \cline{3-4} 
 &  & long & 5532 \\ \hline
\end{tabular}
\label{scenario-cat}\vspace{-0.5cm}
\end{table}
We split the Argoverse validation set into $12$ groups according to the criteria discussed in Sec. \ref{eval-proto}. The statistics are reported in Table \ref{scenario-cat}. We empirically set $\alpha_1$, $\alpha_2$, and $\alpha_3$ as 10$\%$, 45$\%$, and 45$\%$, respectively. In the kinematic compliancy test, we set the range of longitudinal acceleration for normal driving as $(-2m/s^2$ - $1.47m/s^2)$. For the scenario extraction, we set $\beta$, which determines whether trajectories are short or long, as $28.8m$.

\subsection{Accuracy, Diversity, and Admissibility}
\label{sec:acc_div_admis}
\newcolumntype{g}{>{\columncolor{Gray}}c}
\begin{table*}[t]
\vspace{0.2cm}
\centering
\caption{Results on the Argoverse validation set and the most challenging scenarios.}\label{tbl:overall_perf}
\resizebox{0.9\textwidth}{!}{%
\begin{tabular}{c|l|cc|cccgg|ccg}
\hline
& & \multicolumn{2}{c|}{Accuracy} & \multicolumn{5}{c|}{Diversity} & \multicolumn{3}{c}{Admissibility} \\
&Models & minFDE↓ & minADE↓ & RF↑ & minFSD↑ & minASD↑ & AAE↑ & AMV↑ & DAO↑ & DAC↑ & ATT↑ \\ \hline
 \parbox[t]{2mm}{\multirow{5}{*}{\rotatebox[origin=c]{90}{\scalebox{0.8}{Overall} }}} &TNT \cite{zhao2021tnt} & 1.73 & 0.95 & 3.98 & \textbf{10.83} & \textbf{2.79} & \underline{13.77} & 6.50 & \textbf{86.53} & 0.9906 & \textbf{0.861} \\
&LaneGCN \cite{liang2020learning} & \underline{1.08} & \underline{0.71} & \underline{4.41} & 3.79 & 0.95 & 11.43 & 6.64 & 73.82 & \underline{0.9917} & \underline{0.830} \\
&HiVT \cite{zhou2022hivt} & \textbf{0.96} & \textbf{0.66} & 3.64 & 0.40 & 0.12 & 9.77 & 6.17 & 70.18 & \textbf{0.9919} & 0.812 \\
&FTGN \cite{aydemir2022trajectory} & \underline{1.08} & 0.73 & 4.14 & 2.51 & 0.68 & \textbf{14.23} & \underline{6.90} & 72.43 & 0.9908 & 0.816 \\
&MMTrans.\cite{huang2022multi} & \underline{1.08} & \underline{0.71} & \textbf{4.64} & \underline{4.51} & \underline{1.14} & 12.14 & \textbf{7.03} & \underline{74.58} & 0.9902 & 0.821 \\ \hline  \hline 

\parbox[t]{2mm}{\multirow{5}{*}{\rotatebox[origin=c]{90}{\scalebox{0.8}{Challenging} }}} &TNT \cite{zhao2021tnt} & 7.38 & 3.54 & 1.94 & 10.90 & 2.88 & \underline{8.63} & 7.05 & \textbf{79.26} & 0.9634 & \textbf{0.798} \\
&LaneGCN \cite{liang2020learning} & 4.51 & 2.40 & \underline{2.68} & \underline{13.48} & \underline{3.46} & 6.73 & 8.90 & 62.53 & \underline{0.9662} & \underline{0.734} \\
&HiVT \cite{zhou2022hivt} & \underline{4.34} & \textbf{2.20} & 2.61 & 1.35 & 0.41 & 5.31 & 8.59 & 60.20 & 0.9655 & 0.706 \\
&FTGN \cite{aydemir2022trajectory} & 4.80 & 2.46 & 2.62 & 7.81 & 2.25 & \textbf{12.05} & \textbf{9.66} & 58.29 & 0.9552 & 0.709 \\
&MMTrans.\cite{huang2022multi} & \textbf{4.30} & \underline{2.30} & \textbf{2.93} & \textbf{16.10} & \textbf{4.12} & 7.35 & \underline{9.35} & \underline{65.10} & \textbf{0.9729} & 0.731 \\ \hline
\end{tabular}%
}
\end{table*}

We begin by providing an overview of the models and their rankings using the existing and proposed metrics. We report the results over the entire validation set, marked as overall and a challenging subset of the data comprising of hardest scenarios, i.e. hard turns with long trajectories. The results are summarised in Table \ref{tbl:overall_perf}. Note that the proposed metrics in the case of overall are weighted according to the degree of difficulty of samples to provide a more representative overview of the performance.

Our first observation of the results is that the best model in terms of accuracy, HiVT, is not the most diverse, nor admissible one. In fact, on RF and DAO, this model ranks last. TNT, on the other hand, exhibits an opposite behavior by ranking high on diversity and admissibility while having the lowest accuracy.

The proposed diversity metrics show an alternative ranking. Thanks to removing length bias, we can see that MMTransformer achieves the highest magnitude diversity. This new ranking suggests that TNT tends to generate longer trajectories, compared to other models, hence being favored in diversity metrics in general. This model, however, achieves a second best score in terms of angular diversity owing to target selection based on road topology.

For admissibility, the top models' rankings are similar based on DAO and ATT. This is generally expected as TNT tends to constrain trajectory endpoints based on center-lines of the lanes, hence producing more compliant trajectories. For other models, the changes in ranking suggest that they are not successful at one or more of the tests proposed as part of ATT (more on this in Sec. \ref{sec:abl_test}).

In challenging scenarios, we see a significant degradation across all metrics for all models. The degree of change, however, is different. For instance, LaneGCN and FTGN that were sharing the same spot with MMTransformer, have more performance degradation compared to this model. The relative differences across other metrics remain stable, with the exception of TNT whose performance on the existing diversity metrics drops drastically due to significantly worse accuracy and HiVT which changes its first position in DAC with MMTransformer. Despite being the most accurate model, HiVT continues being at the bottom of the diversity and admissibility rankings across all metrics.

\begin{figure}[t]
\centering
\begin{subfigure}{}\includegraphics[width=0.9\columnwidth, trim={0.7cm 0.5cm 1.1cm 1.1cm},clip]{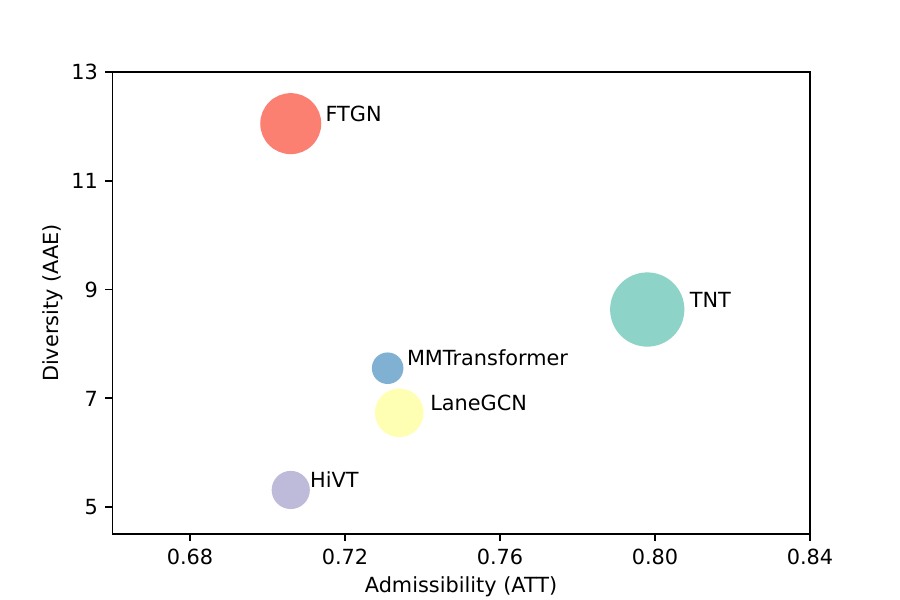} \end{subfigure} \\ 
\begin{subfigure}{}\includegraphics[width=0.9\columnwidth, trim={0.7cm 0cm 1.1cm 1.1cm},clip]{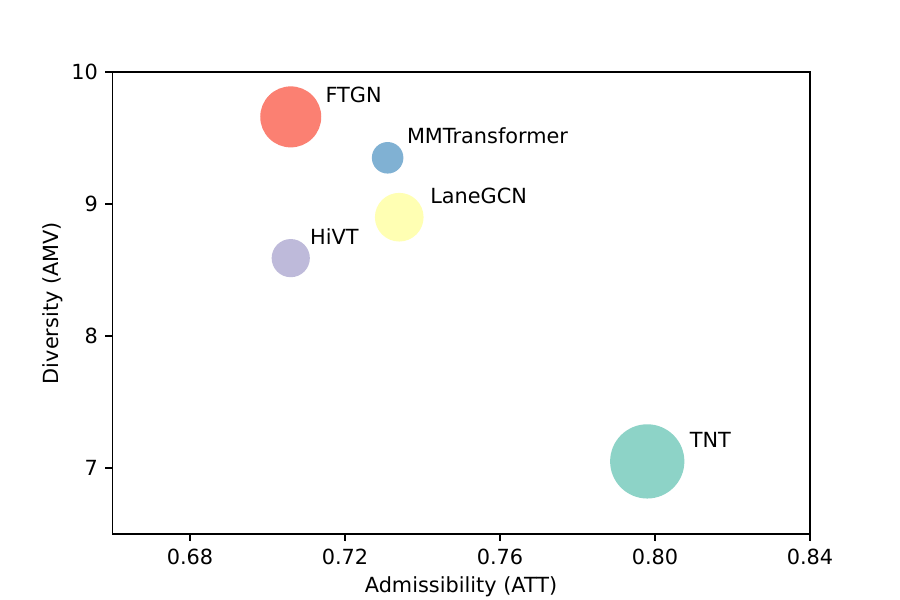}\end{subfigure} \vspace{-2.5mm}
\caption{Performance comparison on the challenging scenarios in terms of diversity, admissibility, and minFDE (represented as the size of circles, so smaller is better).}\label{fig:balance} \vspace{-0.8cm}
\end{figure}

% \begin{figure}
%     \centering
%     \includegraphics[width=5.5cm, trim={0cm 0cm 15cm 0cm},clip]{figures/ranking.pdf}
% \caption{Performance comparison on the challenging scenarios in terms of diversity, admissibility, and minFDE (represented as the size of circles, so the smaller is better).}\label{fig:balance}
% \end{figure}

\subsection{Scenario-based Analysis}
\begin{table*}[]
\vspace{0.2cm}
\centering
\caption{Results on the extracted scenarios with the existing and proposed metrics for diversity and admissibility.} \label{tbl:scenario_analysis}
\resizebox{\textwidth}{!}{
\begin{tabular}{c|l|cccccc|cccccc}
\hline
 &  & \multicolumn{6}{c|}{Cruising} & \multicolumn{6}{c}{Turn} \\
 &  & \multicolumn{3}{c}{Short} & \multicolumn{3}{c|}{Long} & \multicolumn{3}{c}{Short} & \multicolumn{3}{c}{Long} \\ \cline{3-14} 
 & Models & AAE↑ & AMV↑ & \multicolumn{1}{c|}{ATT↑} & AAE↑ & AMV↑ & ATT↑ & AAE↑ & AMV↑ & \multicolumn{1}{c|}{ATT↑} & AAE↑ & AMV↑ & ATT↑ \\ \hline
\parbox[t]{2mm}{\multirow{5}{*}{\rotatebox[origin=c]{90}{\scalebox{0.8}{Hard} }}} & TNT \cite{zhao2021tnt} & 8.71 & \textbf{6.63} & \multicolumn{1}{c|}{\textbf{0.894}} & \underline{1.68} & \textbf{7.75} & 0.885 & \textbf{20.10} & 5.83 & \multicolumn{1}{c|}{\textbf{0.875}} & \underline{8.63} & 7.05 & \textbf{0.798} \\
 & LaneGCN \cite{liang2020learning} & \textbf{9.95} & 5.86 & \multicolumn{1}{c|}{0.842} & 1.52 & 6.80 & 0.883 & 16.95 & 5.86 & \multicolumn{1}{c|}{\underline{0.849}} & 6.73 & 8.90 & \underline{0.734} \\
 & HiVT \cite{zhou2022hivt} & 8.55 & 5.26 & \multicolumn{1}{c|}{0.818} & 1.46 & 5.91 & \textbf{0.902} & 14.87 & 5.52 & \multicolumn{1}{c|}{0.827} & 5.31 & 8.59 & 0.706 \\
 & FTGN \cite{aydemir2022trajectory} & \underline{9.18} & \underline{6.30} & \multicolumn{1}{c|}{0.852} & \textbf{2.53} & \underline{7.00} & \underline{0.891} & \underline{19.45} & \underline{5.91} & \multicolumn{1}{c|}{0.832} & \textbf{12.05} & \textbf{9.66} & 0.709 \\
 & MMTrans.\cite{huang2022multi} & 6.27 & 6.29 & \multicolumn{1}{c|}{\underline{0.862}} & 1.07 & 6.84 & 0.867 & 18.50 & \textbf{6.31} & \multicolumn{1}{c|}{0.834} & 7.35 & \underline{9.35} & 0.730 \\ \hline \hline

\parbox[t]{2mm}{\multirow{5}{*}{\rotatebox[origin=c]{90}{\scalebox{0.8}{Middle} }}} & TNT \cite{zhao2021tnt} & \textbf{7.26} & \textbf{7.29} & \multicolumn{1}{c|}{0.916} & \textbf{1.35} & \textbf{8.10} & 0.921 & \textbf{15.93} & \textbf{6.45} & \multicolumn{1}{c|}{0.884} & \underline{2.72} & \textbf{7.70} & \textbf{0.903} \\
 & LaneGCN \cite{liang2020learning} & 4.66 & 5.15 & \multicolumn{1}{c|}{\underline{0.919}} & 1.15 & \underline{5.73} & 0.935 & 9.58 & 5.07 & \multicolumn{1}{c|}{\textbf{0.902}} & 1.84 & 6.31 & 0.892 \\
 & HiVT \cite{zhou2022hivt} & 2.65 & 4.22 & \multicolumn{1}{c|}{0.911} & 0.45 & 4.32 & \textbf{0.958} & 6.94 & 4.44 & \multicolumn{1}{c|}{\underline{0.900}} & 1.04 & 5.18 & 0.897 \\
 & FTGN \cite{aydemir2022trajectory} & \underline{5.84} & \underline{5.32} & \multicolumn{1}{c|}{0.914} & \underline{1.32} & 5.61 & \underline{0.940} & 11.71 & 5.10 & \multicolumn{1}{c|}{0.894} & \textbf{3.69} & \underline{6.36} & 0.897 \\
 & MMTran.\cite{huang2022multi} & 3.36 & 5.61 & \multicolumn{1}{c|}{\textbf{0.927}} & 0.62 & 5.60 & 0.923 & \underline{10.89} & \underline{5.45} & \multicolumn{1}{c|}{0.886} & 1.50 & 6.19 & \underline{0.898} \\ \hline \hline
\parbox[t]{2mm}{\multirow{5}{*}{\rotatebox[origin=c]{90}{\scalebox{0.8}{Easy} }}} & TNT \cite{zhao2021tnt} & \textbf{9.09} & \textbf{7.77} & \multicolumn{1}{c|}{0.944} & \textbf{1.18} & \textbf{8.37} & 0.940 & \textbf{10.53} & \textbf{6.83} & \multicolumn{1}{c|}{0.908} & \underline{1.73} & \textbf{8.23} & 0.936 \\
 & LaneGCN \cite{liang2020learning} & 3.36 & 4.15 & \multicolumn{1}{c|}{0.969} & 0.87 & \underline{5.08} & 0.958 & 4.63 & 4.42 & \multicolumn{1}{c|}{\underline{0.935}} & 1.13 & \underline{5.22} & 0.947 \\
 & HiVT \cite{zhou2022hivt} & 1.87 & 2.76 & \multicolumn{1}{c|}{\textbf{0.971}} & 0.22 & 3.36 & \textbf{0.985} & 2.64 & 3.44 & \multicolumn{1}{c|}{\textbf{0.937}} & 0.41 & 3.57 & \textbf{0.972} \\
 & FTGN \cite{aydemir2022trajectory} & \underline{4.78} & 4.09 & \multicolumn{1}{c|}{0.959} & \underline{0.88} & 4.82 & \underline{0.967} & \underline{6.79} & 4.44 & \multicolumn{1}{c|}{0.933} & \textbf{2.03} & 4.85 & \underline{0.964} \\
 & MMTran.\cite{huang2022multi} & 3.61 & \underline{4.71} & \multicolumn{1}{c|}{\underline{0.970}} & 0.48 & 4.99 & 0.941 & 5.02 & \underline{4.86} & \multicolumn{1}{c|}{0.931} & 0.76 & 4.95 & 0.947 \\ \hline
\end{tabular}%
}
\end{table*}

\begin{figure*}
     \centering
     \begin{subfigure}{}
         \centering
         \includegraphics[width=0.24\textwidth]{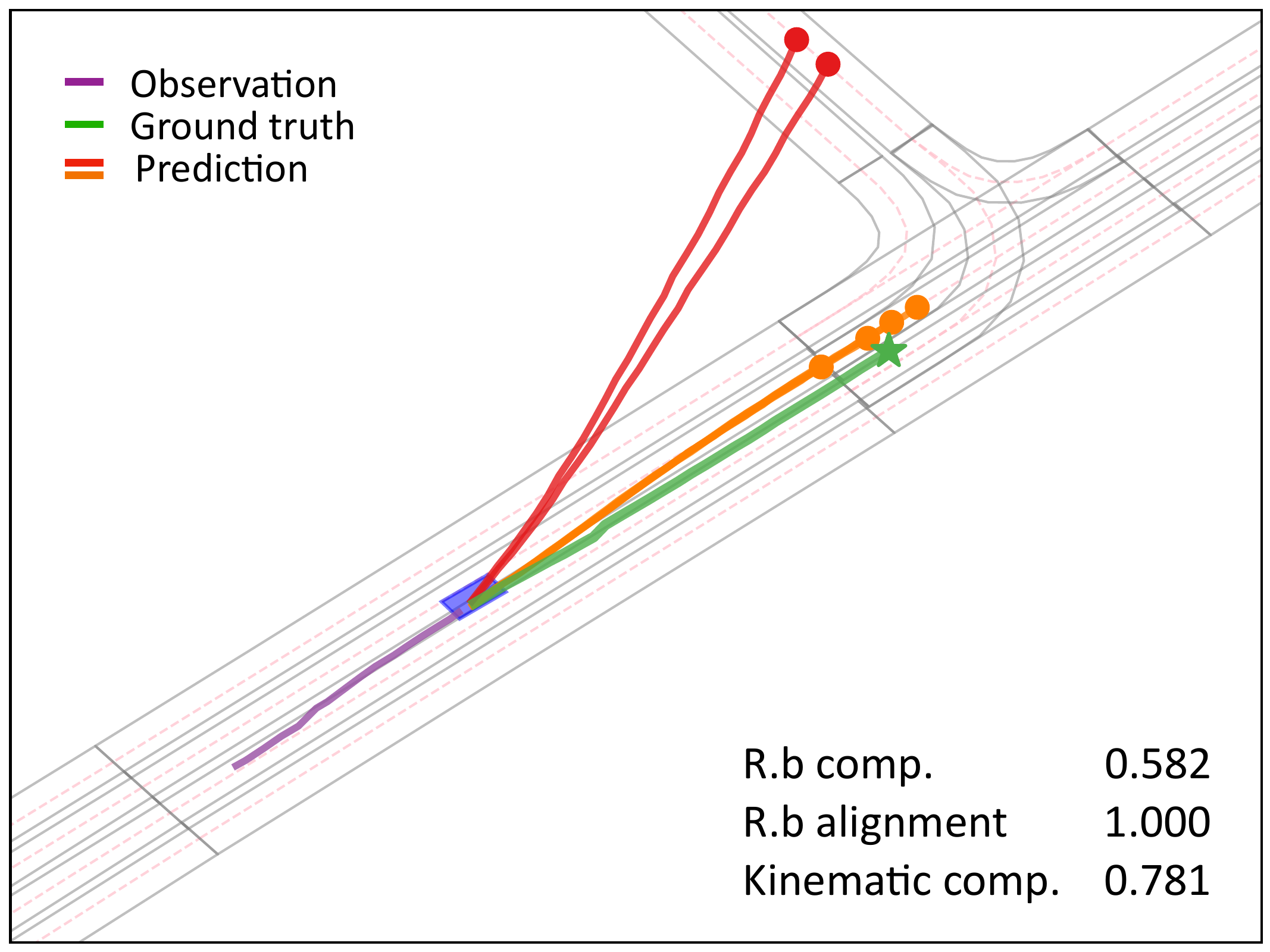}
     \end{subfigure}
      \hspace{-4mm}
      \begin{subfigure}{}
         \centering
         \includegraphics[width=0.24\textwidth]{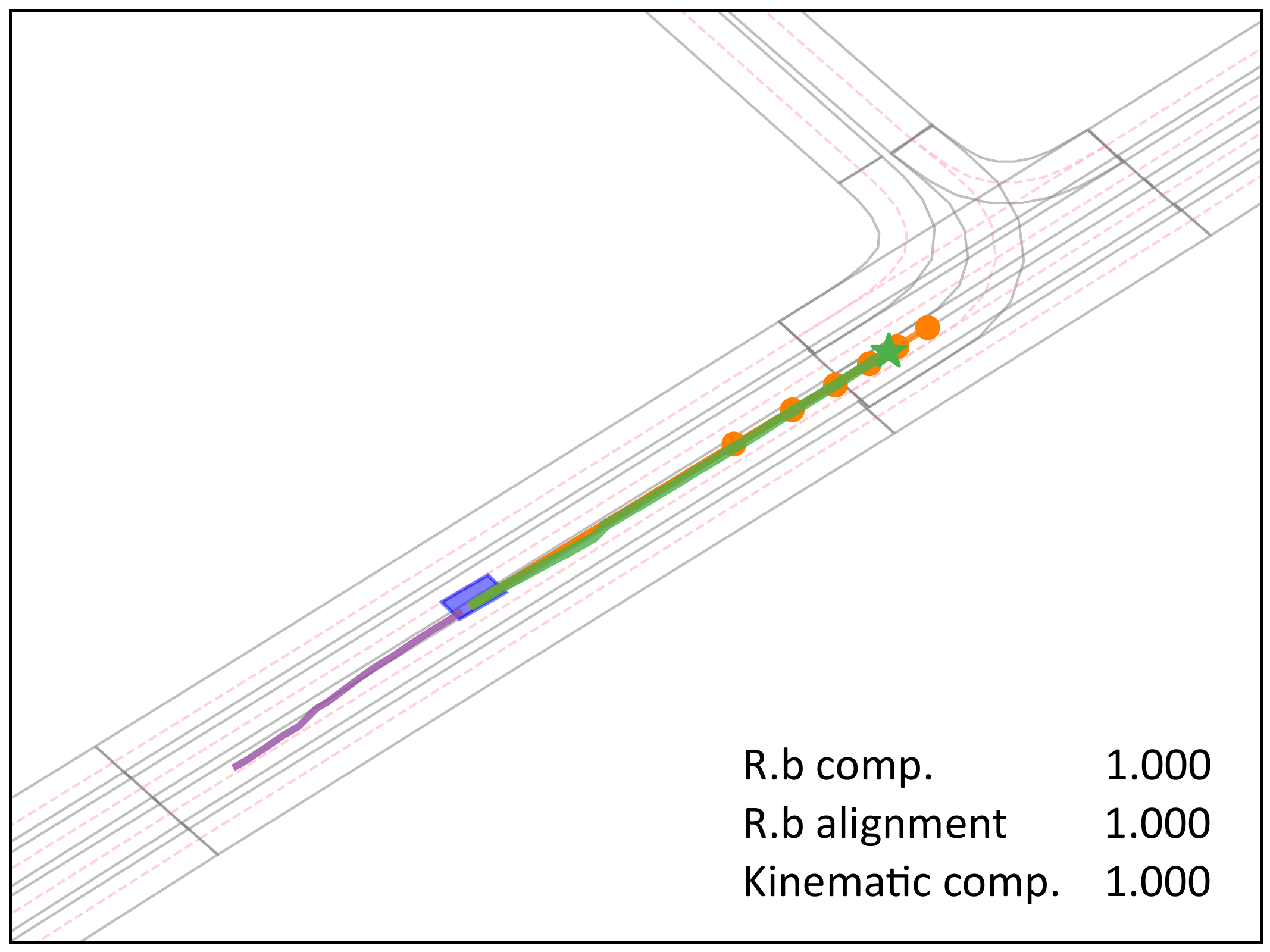}
         \label{fig:five over x}
     \end{subfigure}\vspace{-0.3cm}
     \hspace{-4mm}
      \begin{subfigure}{}
         \centering
         \includegraphics[width=0.24\textwidth]{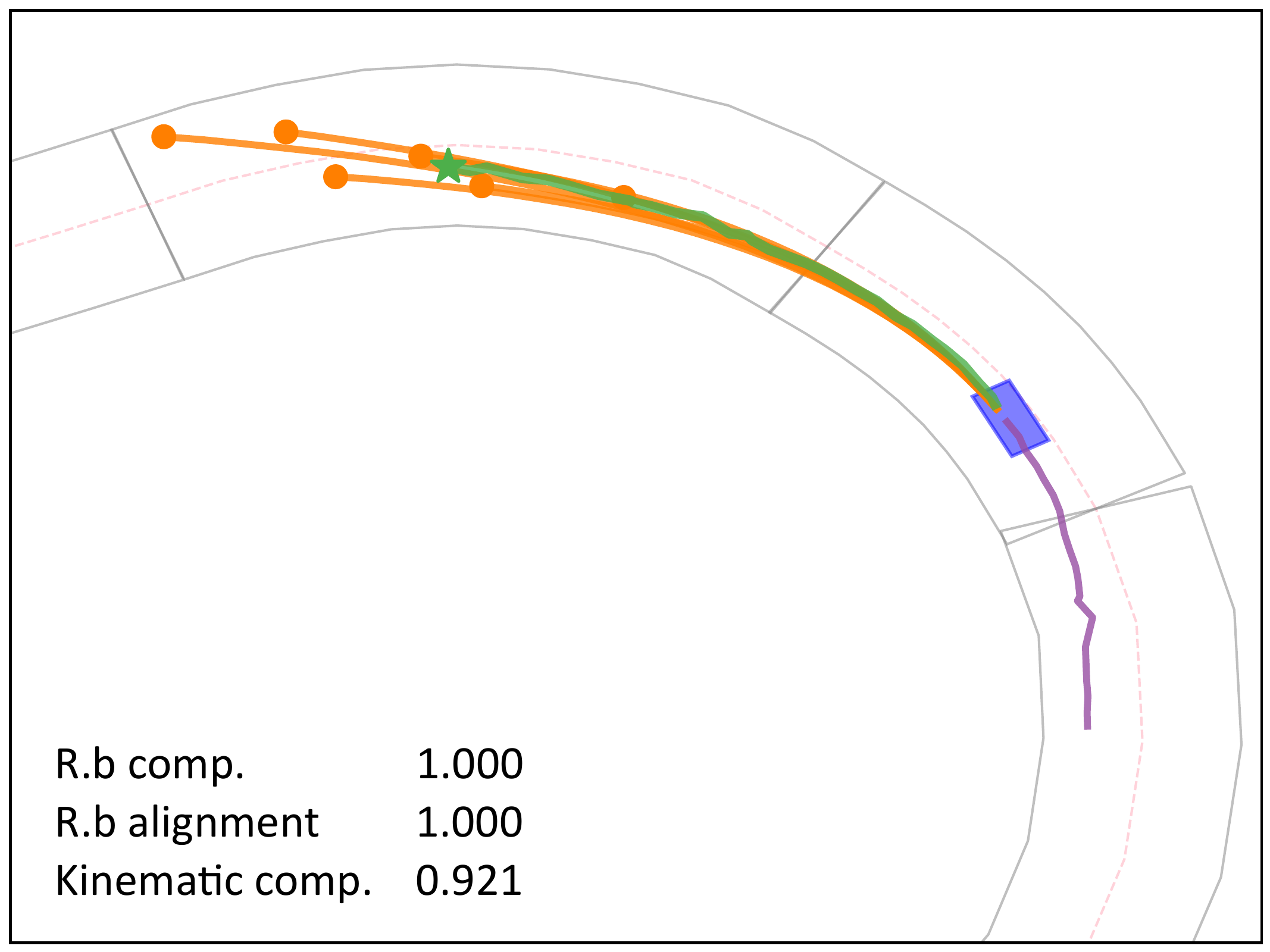}
     \end{subfigure}     \hspace{-4mm}
     \begin{subfigure}{}
         \centering
         \includegraphics[width=0.24\textwidth]{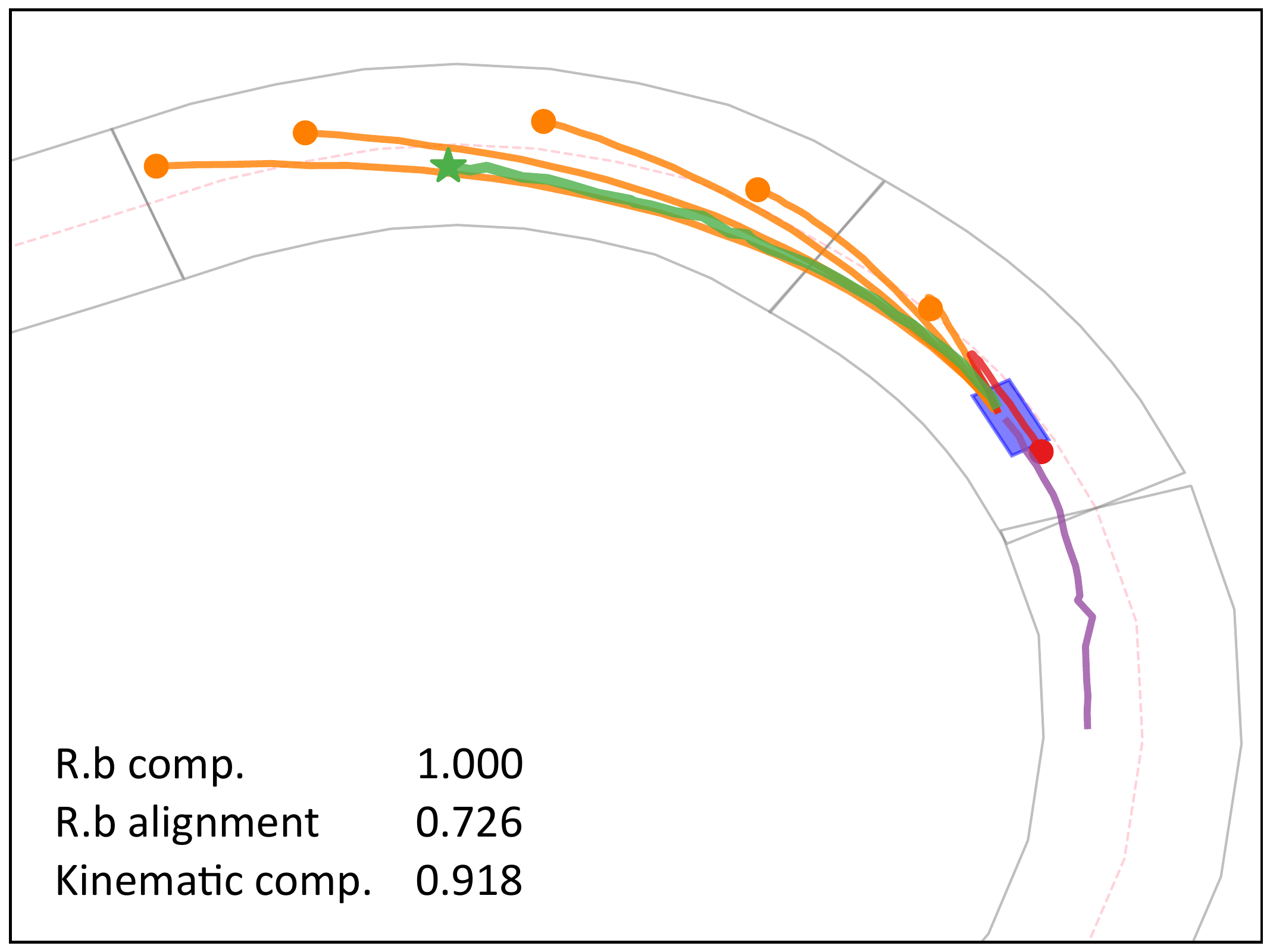}
     \end{subfigure}
         \hspace{-3mm}
        \caption{Qualitative samples of different ATT test scores for predicted trajectories. The red lines indicate inadmissible trajectories. From left to right, the samples are generated by TNT, MMTransformer, TNT and LaneGCN.}
        \label{fig:ablation-fi}\vspace{-0.3cm}
\end{figure*}

To get a better sense of performance differences, we take a look at a finer breakdown of scenarios. For this purpose, we use the criteria described in Sec. \ref{scenario-cat} and split the samples into 12 categories (see Table \ref{scenario-cat} for statistics of each category). For each model under a given scenario, we report the results using the proposed metrics.

As shown in Table \ref{tbl:scenario_analysis}, TNT appears to be a more successful method in terms of diversity, however, with a different degrees of success. In general, in hard turn scenarios, TNT ranks second last and last in AMV, suggesting that this model does not effectively model vehicles' dynamics.

The biggest change in the ranking can be seen in terms of the admissibility metric, ATT. HiVT maintains the first place throughout easy scenarios and performs reasonably in middle scenarios. However, in Hard scenarios, HiVT is placed last in all cases except cruising long where the performance gap between all models is relatively smaller compared to the gaps in metrics in other scenarios. TNT's admissibility generally tends to be lagging in cruising cases, especially with longer horizons. The fluctuations in the performance of the models highlight their limitations under different conditions and the impact of certain scenarios' properties on their performance. 

\subsection{How to Select the Best Model}

As discussed earlier, when evaluating the performance of trajectory prediction models, there are three considerations: trajectories should be accurate, diverse and admissible. In Sec. \ref{sec:acc_div_admis}, we showed that performing good based on one class of metrics does not necessarily translate to a good performance using another class of metrics. This raises the question of how one should select (or design) a model based on the performance? The answer to this question is not intuitive and highly depends on the use case of the model which may put more emphasis on one aspect of the performance, e.g. diversity, over the other, e.g. accuracy. However, for ranking models based on multiple criteria, one can look at how balanced the performance of the models are. For this purpose, we report the results using a three-dimensional representation as illustrated in Fig. \ref{fig:balance}. 

We can see that when considering challenging cases, at two ends of the spectrum, TNT tends to be most successful in terms of admissibility and FTGN in terms of diversity. However, taking into account the accuracy of these models, both tend to lag behind others. Considering all three metrics, MMTransformer and LaneGCN offer a more balanced performance. These models are placed generally in the center areas between diversity and admissibility axes and at the same time have reasonable accuracy. 

\subsection{Ablation study on Contributions of Tests in ATT metric}\label{sec:abl_test}
\begin{table}[]
\centering
\caption{Contributions of the tests in ATT. R.b and comp. stand for road boundary and compliancy, respectively.}
\label{tbl:abl_components}
\begin{tabular}{l|ccc}
\hline
models & R.b comp. & Kinematic comp. & R.b alignment \\ \hline
TNT \cite{zhao2021tnt} & \underline{0.938} & 0.918 & \textbf{0.822} \\
LaneGCN \cite{liang2020learning} & 0.832 & \underline{0.928} & 0.542 \\
HiVT \cite{zhou2022hivt} & 0.807 & \textbf{0.935} & 0.546 \\
FTGN \cite{aydemir2022trajectory} & 0.916 & 0.927 & 0.561 \\
MMTrans.\cite{huang2022multi} & \textbf{0.977} & 0.922 & \underline{0.628} \\ \hline
\end{tabular} \vspace{-0.5cm}
\end{table}
We examine the models in terms of individual ATT's components, namely road boundary, kinematic, and road boundary alignment compliancy tests. As shown in Table \ref{tbl:abl_components}, the models have different strengths. MMTransformer stands out in terms of road boundary compliancy while HiVT is best in kinematic and TNT in road alignment compliancy tests. In general, road alignment is the weakest point of all models where there is a large gap between TNT and the rest. TNT's superior performance can be attributed to its target selection mechanism which constrains the predicted trajectories to correct lane center-lines.

In terms of other tests, while all models are fairly kinematically compliant, they are not equally successful in road boundary test. In fact, we can see a big difference between the bottom two models, LaneGCN and HiVT and the top two, TNT and MMTransfomer. Such differences can show the effectiveness of map encoding methods used by different models. We illustrate examples of generated trajectories by the models and corresponding tests scores in Fig. \ref{fig:ablation-fi}.

\section{Conclusions}
In this paper, we presented a new paradigm for evaluating the performance of vehicle trajectory prediction models for autonomous driving. Our proposed approach consists of a new method to extract scenarios from existing autonomous driving datasets based on characteristics of the data as well as performance agreement among prediction models. Additionally, we proposed three new metrics for measuring diversity and admissibility of trajectories by eliminating length and accuracy biases in the existing metrics.

We conducted extensive set of experiments using the proposed evaluation approach on state-of-the-art trajectory prediction models and provided an alternative perspective of their performance. We showed that good performance under one set of criteria does not necessarily translate to good performance across all scenarios or metrics. Additionally, we demonstrated how the newly proposed metrics can be used to characterize the performance of the models. For future work, we will consider a finer evaluation at model architectural level by conducting studies on models' design choices using the proposed CRITERIA evaluation protocol.

\bibliographystyle{IEEEtran}
\bibliography{ref}

\end{document}